\documentclass[letterpaper, 10 pt, conference]{ieeeconf}
\IEEEoverridecommandlockouts
\usepackage{cite}
\usepackage{pdfpages}
\usepackage{amsmath,amssymb,amsfonts}
\usepackage{algorithmic}
\usepackage{graphicx}
\usepackage{textcomp}
\usepackage{xcolor}
\usepackage{multicol}
\usepackage{hyperref}
\usepackage{subfig}
\usepackage{placeins}
\usepackage{flushend}

\emergencystretch=\maxdimen
\def\BibTeX{{\rm B\kern-.05em{\sc i\kern-.025em b}\kern-.08em
    T\kern-.1667em\lower.7ex\hbox{E}\kern-.125emX}}

\title{``One Soy Latte for Daniel'': Visual and Movement Communication of Intention from a Robot Waiter to a Group of Customers}

\author{Seung Chan Hong$^{1}$, Leimin Tian$^{1}$, Akansel Cosgun$^{2}$ and Dana Kuli\'c$^{1}$
\thanks{$^{1}$Seung Chan Hong, Leimin Tian and Dana Kuli\'c are with the Faculty of Engineering,
        Monash University, Melbourne, VIC 3800, Australia
        {\tt\small shon0019@student.monash.edu}, {\tt\small Leimin.Tian@monash.edu},  {\tt\small dana.kulic@monash.edu}}%
\thanks{$^{2}$Akansel Cosgun is with the School of Information Technology, Deakin University
        {\tt\small Akansel.Cosgun@deakin.edu.au}}%
\thanks{D. Kuli{\'c} is supported by the ARC Future Fellowship (FT200100761).}
}
\begin{document}

\maketitle

\begin{abstract}
Service robots are increasingly employed in the hospitality industry for delivering food orders in restaurants. However, in current practice the robot often arrives at a fixed location for each table when delivering orders to different patrons in the same dining group, thus requiring a human staff member or the customers themselves to identify and retrieve each order. This study investigates how to improve the robot's service behaviours to facilitate clear intention communication to a group of users, thus achieving accurate delivery and positive user experiences. 
Specifically, we conduct user studies (N=30) with a Temi service robot as a representative delivery robot currently adopted in restaurants. We investigated two factors in the robot's intent communication, namely visualisation and movement trajectories, and their influence on the objective and subjective interaction outcomes. A robot personalising its movement trajectory and stopping location in addition to displaying a visualisation of the order yields more accurate intent communication and successful order delivery, as well as more positive user perception towards the robot and its service. Our results also showed that individuals in a group have different interaction experiences. 
\end{abstract}

\section{Introduction}

There has been an increasing deployment of social robots for in-restaurant services, such as taking and delivering orders, to address labour shortages and enhance customer experience~\cite{app112210702, byrd2021robot}. Dining is often a social experience. Robots interacting with users in such settings need to address both objective service outcomes and subjective user experiences. Existing implementations of order delivery robots have focused on reliable navigation and collision avoidance~\cite{bai2021research, jain2023does}. However, the order delivery stage in which a robot directly interacts with users often follows scripted behaviours with the same stopping location used for each table. A group of users seated at the same table is considered as an integral entity while possible differences in individual users' experience are overlooked. Thus, the robot has limited capability to communicate its service intention to individual users and often requires supervision or intervention from human staff to complete the delivery. This can negatively impact service efficiency and user satisfaction~\cite{byrd2021robot}. 

\begin{figure}[tb]
    \centering
    \includegraphics[width=\linewidth]{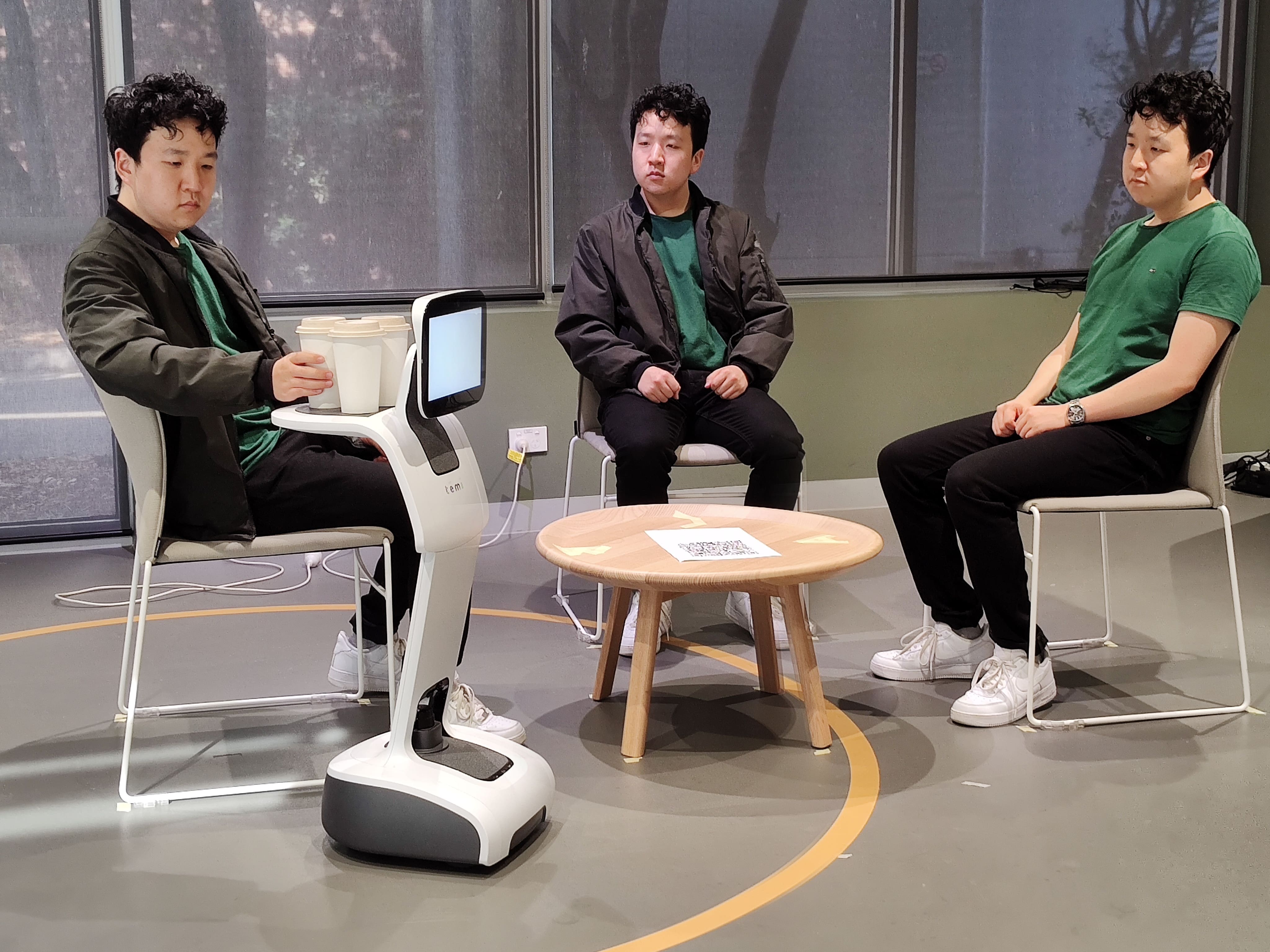}\label{fig:teaser}
     \caption{We study the effect of stopping location and visual indicators for a restaurant delivery scenario}
\label{fig:teaser}
\end{figure}

We are motivated to investigate a service robot's intent communication when delivering orders to a group of users, as demonstrated in Figure~\ref{fig:teaser}. We aim to understand how different order delivery behaviours influence the robot's accuracy in communicating its intent to each human receiver in the group, as well as individual users' subjective perception towards the robot and the interaction. To the best of our knowledge, this is the first study on robot intent communication in group service interaction using both visual and movement cues. We contribute to current research on social robots and robot-group interaction by demonstrating the benefits of combining visualisation and motion. Our study also informs how restaurant service robots can enhance customers' experiences by providing intuitive and enjoyable interaction. 

Inspired by literature on human-robot interaction (HRI) and object handovers~\cite{ortenzi2021object}, we focus on two aspects of the robot's order delivery behaviours, namely motion and visualisation. We investigate the influence of customising the robot's movement trajectory and stopping location to individual users in a group as opposed to the existing approach of using one location for the whole table. In addition, we study the influence of displaying visual information related to the orders as opposed to no visualisation. We conduct user studies with a Temi robot, a wheeled robot with a tablet and a back tray, as a representative of delivery robots currently deployed in restaurant and hospitality applications. We measured the objective outcomes as accuracy in intent communication, namely if a user was able to retrieve the correct orders intended for them on the first attempt or not. Further, we used questionnaires to evaluate the subjective outcomes in terms of a user's perception towards the robot and the interaction. Video recordings of the user study and questionnaire responses have been provided as a publicly available dataset to support future research on group interaction with robots in the service context.\footnote{Dataset: \url{https://doi.org/10.26180/25441144.v2}}

\section{Related works}

\subsection{Motion and physical proximity in HRI}\label{subsec:bg-motion}
A service robot's movement trajectories when approaching a user and its orientation when stopping and interacting with the user play a significant role in how humans perceive and interact with it. Current research suggests that people prefer a robot to approach them from their left or right side while approaching directly from the front can be perceived as uncomfortable, even threatening~\cite{10.1145/1121241.1121272}. In alignment with current research, we design the Temi robot to approach customers from their sides at the time of order delivery. 

Furthermore, existing works suggest that incorporating the socio-spatio-temporal characteristics of human perception in a robot's behavioural designs enhances its perceived comfort and safety in HRI~\cite{8036225}. Understanding the social implications of physical proximity informs the design of socially appropriate robot motions~\cite{7832481}. Following this previous work, we designed the stopping location of the robot to be at a suitable distance, namely at arm's length so that customers can easily reach and collect their orders, while at the same time not being too close to cause discomfort. In addition, the robot's pose can influence whether or not a mobile robot is perceived as socially acceptable~\cite{9775638}. Thus, we implement the order delivery behaviour sequence to first have the robot's screen facing the participant, giving a spoken confirmation ``Your order is here'' while displaying visualisation if relevant. The robot then rotates 180 degrees to have its tray facing the participant (see Figure~\ref{fig:teaser}), which allows comfortable and accessible order collection.

\subsection{Visual display for intent communication}\label{subsec:bg-vis}
Existing studies have shown that people complete human-robot collaboration tasks faster when they are guided by visual instructions, such as screen-based visualisation~\cite{8359206} or a combination of static and dynamic visual cues~\cite{10342222}. 
Previous research suggests that compared to dynamic visual cues, such as videos or animations, a simple static image can be sufficient and even more effective in conveying a robot's intention to humans~\cite{fernandez2018light}. Thus, in this work, we designed static and concise visualisations which are displayed on the robot's screen. Previous research suggested that humans can interpret the intentions of a robot solely relying on visual cues, even in their first interaction with the robot~\cite{6602436}. However, the robots are often stationary in these studies and engage in a one-on-one HRI. It is unclear whether or not combining visual and movement cues can bring additional benefits to the objective and subjective outcomes of human-robot collaboration, especially in the group interaction scenario. Thus, we are motivated to investigate how visual and movement cues on their own and in combination will influence a robot's intention communication effectiveness when interacting with a group of users.

\subsection{Restaurant delivery robots}\label{subsec:bg-delivery}
Order delivery in restaurants is a common application of service robots. Previous work has shown that such robots' perceived competence and warmth were positively correlated with the customers' positive emotions towards the robots and their service satisfaction~\cite{CHEN2023103482}. In addition, other studies have found that the perception of innovativeness evoked by robot deployment enhanced customers' perceived service quality and their restaurant experience~\cite{cha2020customers}. Previous research has identified four key factors contributing to a customer's perception towards a robot's service quality, namely automation, personalisation, efficiency, and precision~\cite{shah2023influence}. This work contributes to the advancement of restaurant delivery robots focusing on the personalisation and precision aspects, by investigating the impact of visual and movement service intent communication behaviours in a group dining scenario. 

Another study investigated robot waiter delivery design~\cite{10.1145/3610977.3634978}. Customers' action of order collection or their feelings of obligation to do so was found to be positively correlated with human or robot-initiated delivery, while robot waiters that initiated the delivery were found to be more noticeable. Our work investigates the effects visualisations have in conjunction with a robot waiter's delivery path. Further, we measure the delivery accuracy of items that are not immediately differentiable from appearance (i.e., coffee cups with similar appearance as opposed to cupcakes with distinct appearance used in~\cite{10.1145/3610977.3634978}).

\section{Methodology}
Given previous research on service robots for food delivery, this paper seeks to extend the existing knowledge by investigating a robot's motion, physical proximity, and visualisations for intent communications when delivering orders to a group of customers. We measure objective service outcomes as the delivery accuracy of items with similar appearance, as well as subjective user experience and service satisfaction in a user study.
The study protocols were reviewed and approved by the Monash University Human Research Ethics Committee (Project ID 41261). 

\subsection{Hypotheses}

Our main research question is ``how can a social robot communicate its intention when delivering orders to a group of customers naturally and effectively?'' Specifically, we investigate the influence of the robot's visual display and stopping location during such service interaction. Our hypotheses are:

\begin{enumerate}
    \item A robot displaying \textbf{visualisation} of the order compared to not displaying visualisation will result in more \textbf{accurate} order delivery and more \textbf{positive} user perception towards the robot and its service
    \item A robot using \textbf{personalised} stopping locations to deliver each customer's order compared to delivering the orders at a general stopping location for the whole group will result in more \textbf{accurate} order delivery and more \textbf{positive} user perception towards the robot and its service
    \item Each \textbf{customer} in the group will have a different perception towards the robot and its service
\end{enumerate}

\subsection{Interaction scenarios}
\begin{figure}[h!]
    \centering
    \subfloat[][General + no visual]{\includegraphics[height=4.5cm]{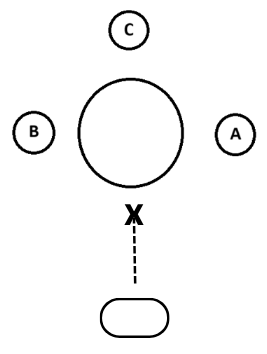}\label{fig:gen_novis}}\quad
    \subfloat[][General + visual]{\includegraphics[height=4.5cm]{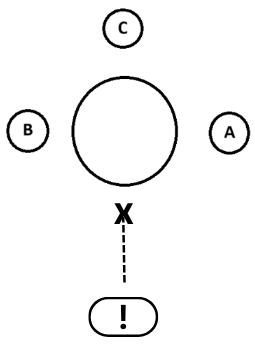}\label{fig:gen_vis}}\\
    
    \subfloat[][Personal + no visual]{\includegraphics[height=4.5cm]{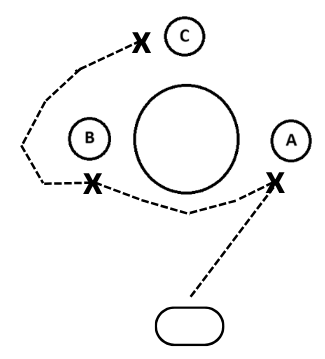}\label{fig:per_novis}}\quad
    \subfloat[][Personal + visual]{\includegraphics[height=4.5cm]{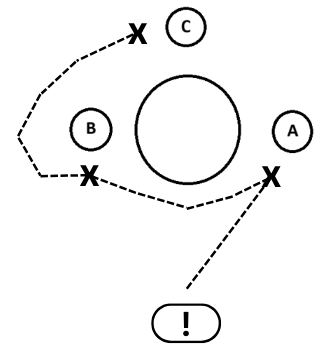}\label{fig:per_vis}}\\
    \subfloat[][Legends of the layout overviews]{\includegraphics[height=4.5cm]{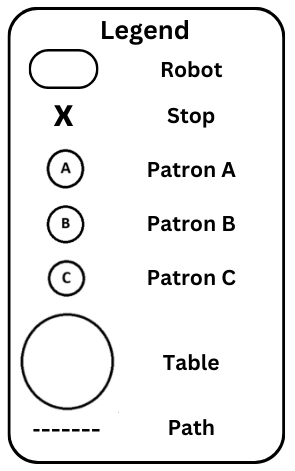}\label{fig:layout_legend}}\quad\quad\quad
    \subfloat[][Temi service robot]{\includegraphics[height=4.5cm]{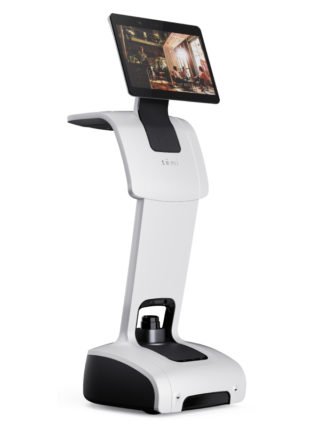}\label{fig:temi}}
     \caption{Each group of three participants experienced four interactions with Temi in random orders. A 2x2 design was used to investigate influence of the robot's visual display (with/without) and stopping location (general/personal) when delivering orders to a group of customers.}
\label{fig:layout}
\end{figure}

\begin{figure}[h!]
    \centering
    \subfloat[][Visualisation for general stop]{\includegraphics[height=3.2cm]{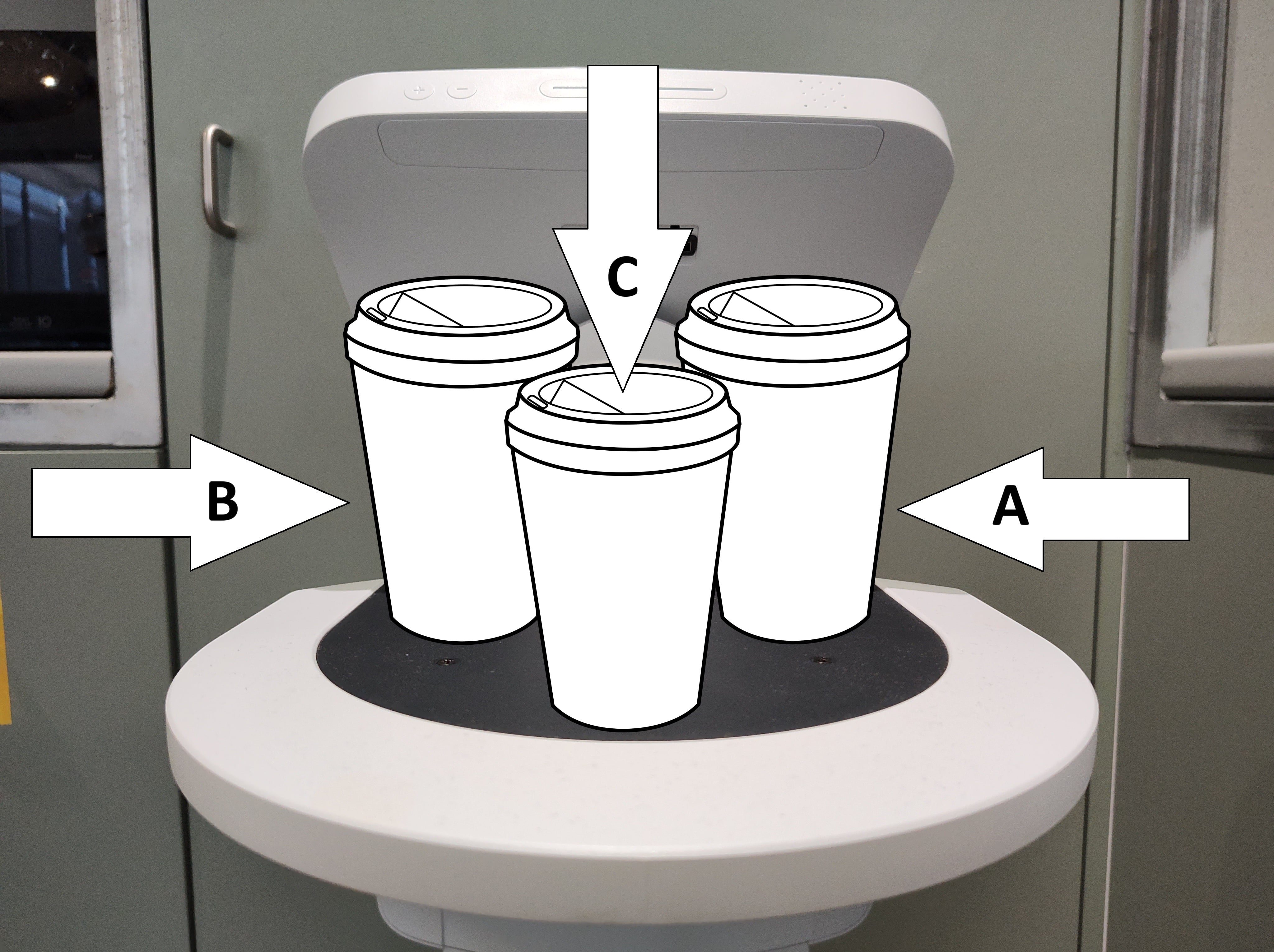}\label{fig:vis_all}}~
    \subfloat[][Visualisation for personal stop]{\includegraphics[height=3.2cm]{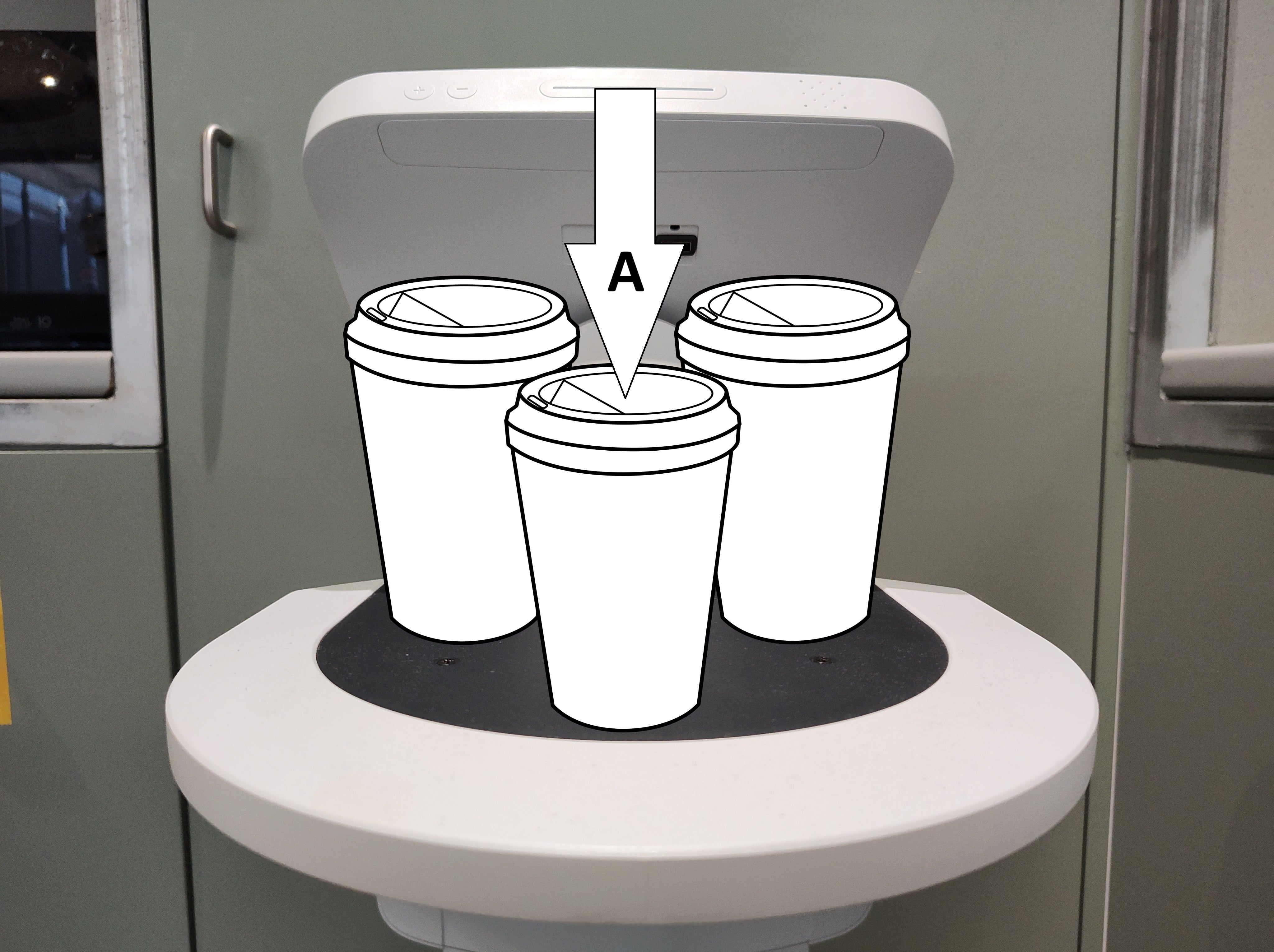}\label{fig:vis_per}}
     \caption{Example delivery visualisation on Temi's tablet}
\label{fig:vis}
\end{figure}

To investigate the above hypotheses, we designed a mixed-model user study, in which participants interacted with a Temi robot in groups of three. As shown in Figure~\ref{fig:layout} and~\ref{fig:vis}, during a session the participants sat at assigned seats as Customers A, B, and C. Each group experienced four interaction scenarios in random order\footnote{Video demonstrations of the four interaction scenarios: \href{https://drive.google.com/file/d/1U-WEWkAaZA5F4KXBRAusoYEEwyn5Rzax/view?usp=sharing}{General + no visual}, \href{https://drive.google.com/file/d/1U6267FJ9DboDS6bsODnBDmrU0KDt7kJJ/view?usp=sharing}{General + visual}, \href{https://drive.google.com/file/d/1UFFuOugiW4QX-80l9eQ0y4MalOIAnoaU/view?usp=sharing}{Personal + no visual}, \href{https://drive.google.com/file/d/1UHbBukqIGs7jG3b7o6ErkEOhyxhhVhaj/view?usp=sharing}{Personal + visual}}:
\begin{itemize}
    \item General stopping location without visualisation
    \item Personal stopping location without visualisation
    \item General stopping location with visualisation
    \item Personal stopping location with visualisation
\end{itemize}


\subsection{Evaluation metrics}
We collected objective and subjective measures to evaluate how intention communication approaches influence the interaction outcomes and user experiences. Specifically, the robot delivered three unmarked paper cups, each containing a treat. Inside the cup at the bottom, a letter is written to indicate if this treat is intended for Customer A, B, or C. The participants were made aware that the location of the cups is changed in between the four interactions and may not correspond to their seating arrangement. Each participant was asked to take one cup from the robot and we recorded order delivery accuracy as whether or not a participant took the correct cup corresponding to their assigned customer role in the first attempt.

An evaluative questionnaire was used to collect participants' subjective perceptions after each interaction session. We used the Godspeed Questionnaire~\cite{bartneck2009measurement} for evaluating perception towards the robot, the human-robot collaboration questionnaire for evaluating the interaction fluency, human-robot trust, and working alliance~\cite{ortenzi2021object}. In addition, we included two ad-hoc questions on service satisfaction and perceived service efficiency. All items were measured on a 5-point Likert scale. We have also performed a manipulation test, where after each interaction session the participants answered if they think the robot made a general delivery to the whole table or a personal delivery (stopping location) and if the robot provided a visual display of the orders or not (visualisation). The questionnaire, participants' responses, and audiovisual recordings of the experiment have been provided in the accompanying dataset.

In addition to quantitative measures, free-text boxes were included in the questionnaire for qualitative evaluation of the conditions. Further, we conducted exit interviews with each participant group and collected their impressions and understanding of the robot and the interaction, as well as comparison to any prior experiences.

\section{Results}
We recruited 30 participants from the university's student population (3 x 10 groups, 5 females, 25 males, age 20.1 $\pm$ 1.5). Two participants reported they have interacted with the Temi robot before, while 11 participants have interacted with restaurant order delivery robots before. In the manipulation test, participants had 90.84\% accuracy in identifying the stopping locations correctly and 95.83\% for visualisation, demonstrating that they were able to distinguish the different conditions reliably. 

Here we report key results comparing the interaction sessions in terms of objective intention communication performance measured as order delivery accuracy, as well as subjective user perception and preferences.

\subsection{Order delivery accuracy}
\begin{figure}[htb]
    \centering
    \includegraphics[trim={0 1cm 0 3cm},clip,width=\linewidth]{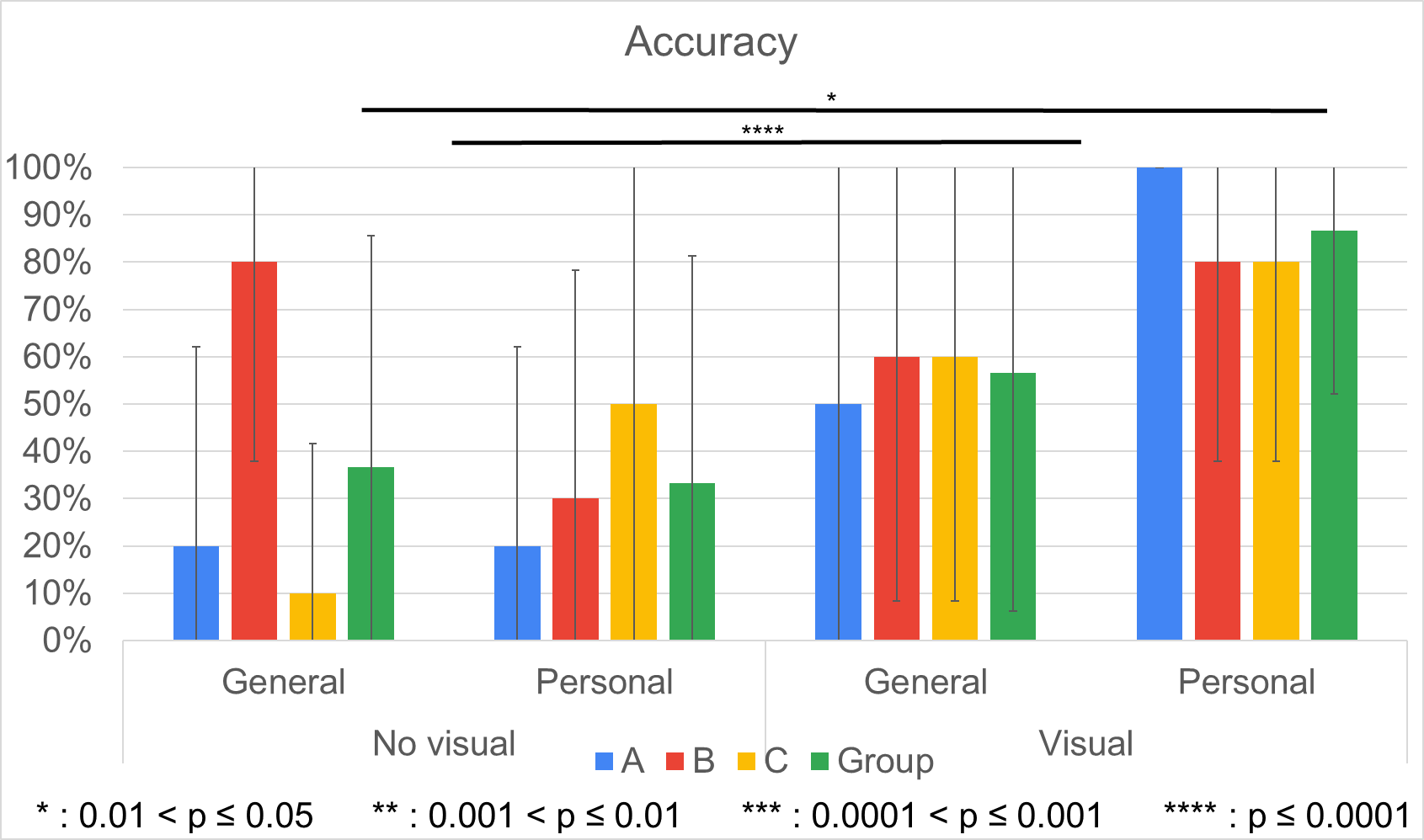}
    \caption{Differences in order delivery accuracy: Combining visualisation and personalised stopping location yielded highest order delivery accuracy. Visualisation, the interaction between visualisation and stopping location, as well as the interaction between stopping location and customer role has a significant influence on order delivery accuracy.}
\label{fig:accuracy}
\end{figure}

In Table~\ref{tab:accuracy} and in Figure~\ref{fig:accuracy}, we report the accuracy of a participant getting the correct order by customer role and for the group as a whole. Note that each participant took one cup for themselves in each interaction session. For group-wise order delivery accuracy, we only consider a delivery successful if all three participants in the group received the correct order.

\begin{table}[htb]
\centering
\caption{Order delivery accuracy by customer role (A/B/C) and group-wise (\%). In the conditions (``Cond.''), ``G'' stands for general stopping location, ``P'' for personal stop, ``V'' for with visualisation, ``N'' for no visualisation.}
\label{tab:accuracy}
\begin{tabular}{|l|r|r|r|r|}
\hline
Cond. & A & B & C & Group\\
\hline
G\texttt{+}N &  20\%&  80\%&  10\%& 37\%\\
\hline
G\texttt{+}V &  50\%&  60\%&  60\%& 57\%\\
\hline
P\texttt{+}N &  20\%&  30\%&  50\%& 33\%\\
\hline
P\texttt{+}V &  \textbf{100\%}&  \textbf{80\%}&  \textbf{80\%}& \textbf{87\%}\\
\hline
\end{tabular}
\end{table}

We conducted a three-way mixed ANOVA, with customer role (A/B/C) as a between-subject factor, stopping location (general/personal) and visualisation (with/without) as within-subject factors and order delivery accuracy as the dependent variable. We found \textbf{visualisation} to significantly influence order delivery accuracy ($F(1,108) = 20.94$, $p << 0.001$) with displaying visualisation leading to higher order delivery accuracy. There was also a significant interaction between \textbf{visualisation} and stopping \textbf{location} ($F(1,108) = 4.33$, $p = 0.04$) with personal stopping resulting in the highest order delivery accuracy when visualisation was displayed. In addition, we found a significant interaction between stopping \textbf{location} and \textbf{customer} role ($F(2,108) = 3.16$, $p = 0.046$). 

As this is the first study that investigates the conjunction of a robot waiter's movement and visualisations in a group dining scenario, we were unable to conduct a power analysis before the experiment. We analysed the effect sizes of our three-way mixed ANOVA calculated as $\eta_{p}^{2}$. \textbf{Visualisation} was found to have a large effect on order delivery accuracy ($\eta_{p}^{2} = 0.16$).

\subsection{Perception towards the robot and the service interaction}
We report the questionnaire results measuring participants' subjective perception towards the robot and the interaction. Before the study, the participants also answered the Godspeed questionnaire to capture their initial perception towards the robot (the ``pre'' rows in Table~\ref{tab:godspeed}). Similar to order delivery accuracy, we performed three-way mixed ANOVA to analyse how stopping location, visualisation, and customer role influence users' subjective perceptions. 

\begin{table}[htb]
\centering
\caption{Perception towards the robot measured by the God speed questionnaire (``Anth.'' is Anthropomorphism, ``Anim.'' is Animacy, ``Like.'' is Likeability, ``Inte.'' is perceived Intelligence, ``Safe.'' is perceived Safety). The ratings are reported as ``mean $\pm$ std'' for all participants and grouped by customer roles. ``pre'' stands for answers in the pre-study questionnaire.}
\label{tab:godspeed}
\begin{tabular}{|l|r|r|r|r|r|}
\hline
\multicolumn{6}{|c|}{\textbf{All participants}}\\
\hline
Cond. & Anth. & Anim. & Like. & Inte. & Safe.\\
\hline
pre & 2.5$\pm$1.0&  2.5$\pm$1.0&  3.6$\pm$0.9&  3.2$\pm$1.0& 3.4$\pm$1.2\\
\hline
G\texttt{+}N &  2.7$\pm$1.2&  2.8$\pm$1.1
&  3.2$\pm$1.1&  2.8$\pm$1.2
& 3.2$\pm$1.2
\\
\hline
P\texttt{+}N&  2.7$\pm$1.1&  2.9$\pm$1.1&  3.3$\pm$1.2&  3.0$\pm$1.2
& 3.2$\pm$1.2
\\
\hline
G\texttt{+}V&  2.8$\pm$1.2&  2.7$\pm$1.1&  3.2$\pm$1.3&  2.9$\pm$1.2
& 3.3$\pm$1.2
\\
\hline
P\texttt{+}V &  \textbf{3.1}$\pm$1.1&  \textbf{3.2}$\pm$1.0
&  \textbf{3.8}$\pm$0.8&  \textbf{3.7}$\pm$0.8
& \textbf{3.6}$\pm$1.1
\\
\hline
\multicolumn{6}{|c|}{\textbf{Customer A}}\\
\hline
Cond. & Anth. & Anim. & Like. & Inte. & Safe.\\
\hline
pre &  2.3$\pm$1.0&  2.4$\pm$1.0&  3.4$\pm$0.8&  3.1$\pm$1.0& 3.2$\pm$1.3\\
\hline
G\texttt{+}N 
&  2.9$\pm$1.1
&  2.8$\pm$1.0
&  3.1$\pm$0.7&  2.6$\pm$1.1
& 3.3$\pm$1.1
\\
\hline
P\texttt{+}N
&  2.9$\pm$1.1&  3.0$\pm$1.2
&  3.3$\pm$1.2&  3.1$\pm$1.3
& 3.6$\pm$1.1
\\
\hline
G\texttt{+}V
&  3.0$\pm$1.2&  2.7$\pm$0.9
&  3.0$\pm$1.1&  2.9$\pm$1.0
& 3.5$\pm$1.1
\\
\hline
P\texttt{+}V &  3.3$\pm$1.2&  3.6$\pm$1.0
&  3.9$\pm$0.9&  3.8$\pm$0.9
& 3.6$\pm$1.2
\\
\hline
\multicolumn{6}{|c|}{\textbf{Customer B}}\\
\hline
Cond. & Anth. & Anim. & Like. & Inte. & Safe.\\
\hline
pre &  2.7$\pm$1.0&  2.6$\pm$1.1&  3.7$\pm$0.9&  3.4$\pm$1.1& 3.6$\pm$1.2\\
\hline
G\texttt{+}N 
&  3.0$\pm$1.1
&  3.1$\pm$1.2
&  3.6$\pm$1.0&  3.4$\pm$0.9
& 3.3$\pm$1.1
\\
\hline
P\texttt{+}N
&  2.7$\pm$0.9
&  2.9$\pm$1.0
&  3.6$\pm$0.6&  3.3$\pm$0.9
& 3.3$\pm$1.1
\\

\hline
G\texttt{+}V
&  3.0$\pm$1.1
&  3.0$\pm$1.1
&  3.6$\pm$1.0&  3.1$\pm$1.0
& 3.4$\pm$1.1
\\
\hline
P\texttt{+}V &  3.0$\pm$1.0
&  3.0$\pm$1.0
&  3.9$\pm$0.8&  3.9$\pm$0.6
& 3.6$\pm$1.0
\\
\hline
\multicolumn{6}{|c|}{\textbf{Customer C}}\\
\hline
Cond. & Anth. & Anim. & Like. & Inte. & Safe.\\
\hline
pre &  2.4$\pm$1.0&  2.5$\pm$1.0&  3.6$\pm$0.9&  3.0$\pm$0.9& 3.3$\pm$1.1\\
\hline
G\texttt{+}N 
&  2.3$\pm$1.1&  2.5$\pm$1.0
&  2.8$\pm$1.3
&  2.5$\pm$1.3
& 3.0$\pm$1.5
\\
\hline
P\texttt{+}N
&  2.5$\pm$1.1
&  2.7$\pm$1.2
&  3.0$\pm$1.5
&  2.7$\pm$1.2
& 2.7$\pm$1.1
\\
\hline
G\texttt{+}V
&  2.4$\pm$1.3&  2.4$\pm$1.2
&  3.1$\pm$1.6
&  2.6$\pm$1.5
& 3.0$\pm$1.3
\\
\hline
P\texttt{+}V &  3.1$\pm$1.1
&  3.0$\pm$1.0&  3.8$\pm$0.8
&  3.6$\pm$0.9
& 3.5$\pm$1.0\\
\hline
\end{tabular}
\end{table}

Regarding subjective perception towards the robot reported in Table~\ref{tab:godspeed}, \textbf{visualisation} showed significant influence on \textit{anthropomorphism} ($F(1,588) = 7.37$, $p = 0.01$), \textit{likeability} ($F = 11.26$, $p << 0.001$), and perceived \textit{intelligence} ($F = 18.02$, $p << 0.001$); Stopping \textbf{location} showed significant influence on \textit{animacy} ($F(1,588) = 11.07$, $p << 0.001$), \textit{likeability} ($F = 22.34$, $p << 0.001$), and perceived \textit{intelligence} ($F = 38.87$, $p << 0.001$); \textbf{Customer} role showed significant influence on \textit{anthropomorphism} ($F(2,588) = 8.94$, $p << 0.001$), \textit{animacy} ($F = 9.07$, $p << 0.001$), \textit{likeability} ($F = 13.52$, $p << 0.001$), perceived \textit{intelligence} ($F = 15.95$, $p << 0.001$), and perceived \textit{safety} ($F = 4.96$, $p = 0.01$). 
In addition, the interaction between \textbf{visualisation} and stopping \textbf{location} showed significant influence on \textit{anthropomorphism} ($F(1,588) = 4.22$, $p = 0.04$), \textit{animacy} ($F = 5.44$, $p = 0.02$), \textit{likeability}($F = 9.08$, $p = 0.003$), and perceived \textit{intelligence} ($F = 14.34$, $p << 0.001$); The interaction between stopping \textbf{location} and \textbf{customer} role was found to have significant influence on \textit{anthropomorphism} ($F(2,588) = 3.51$, $p = 0.03$) and \textit{animacy} ($F = 4.48$, $p = 0.01$). 

\begin{table}[htb]
\centering
\caption{Perception towards the interaction measured as human-robot collaboration fluency (``Fluen.''), trust towards the robot (``Trust''), working alliance (``Alliance''), service satisfaction (``Safis.''), and perceived efficiency of the delivery (``Effec.''). The ratings are reported as ``mean $\pm$ std'' for all participants and grouped by customer roles.}
\label{tab:colab}
\begin{tabular}{|l|r|r|r|r|r|}
\hline
\multicolumn{6}{|c|}{\textbf{All participants}}\\
\hline
Cond. & Fluen. & Trust & Alliance & Satis. & Effec.\\
\hline
G\texttt{+}N 
&  2.8$\pm$1.3
&  2.8$\pm$1.2
&  3.0$\pm$1.2
&  2.7$\pm$1.2
& \textbf{3.8}$\pm$1.2
\\
\hline
P\texttt{+}N
&  3.0$\pm$1.3
&  2.9$\pm$1.2
&  2.8$\pm$1.3
&  3.1$\pm$1.3
& 3.7$\pm$1.1
\\
\hline
G\texttt{+}V
&  2.9$\pm$1.4
&  3.3$\pm$1.4
&  3.2$\pm$1.4
&  2.8$\pm$1.2
& 3.2$\pm$1.0
\\
\hline
P\texttt{+}V &  \textbf{4.0$\pm$0.9}
&  \textbf{3.8}$\pm$1.0
&  \textbf{4.1}$\pm$0.9
&  \textbf{3.9}$\pm$0.9
& 3.5$\pm$1.0
\\
\hline
\multicolumn{6}{|c|}{\textbf{Customer A}}\\
\hline
Cond. & Fluen. & Trust & Alliance & Satis. & Effec.\\
\hline
G\texttt{+}N 
&  3.0$\pm$1.1
&  2.8$\pm$1.0
&  3.0$\pm$1.1
&  2.8$\pm$1.2
& 3.6$\pm$1.1
\\
\hline
P\texttt{+}N
&  3.2$\pm$1.7
&  2.9$\pm$1.3
&  3.1$\pm$1.3
&  3.5$\pm$1.1
& 3.8$\pm$0.9
\\
\hline
G\texttt{+}V
&  3.1$\pm$1.2
&  3.7$\pm$1.3
&  3.2$\pm$1.4
&  2.8$\pm$1.2
& 3.4$\pm$0.8
\\
\hline
P\texttt{+}V &  4.1$\pm$0.9
&  3.8$\pm$1.1
&  4.1$\pm$1.1
&  4.2$\pm$0.9
& 3.5$\pm$0.9
\\
\hline
\multicolumn{6}{|c|}{\textbf{Customer B}}\\
\hline
Cond. & Fluen. & Trust & Alliance & Satis. & Effec.\\
\hline
G\texttt{+}N 
&  3.1$\pm$1.5
&  3.4$\pm$1.4
&  3.8$\pm$1.0
&  3.1$\pm$1.2
& 4.5$\pm$0.5
\\
\hline
P\texttt{+}N
&  3.1$\pm$1.4
&  3.4$\pm$1.2
&  2.7$\pm$1.4
&  3.0$\pm$1.2
& 4.3$\pm$0.7
\\
\hline
G\texttt{+}V
&  3.5$\pm$1.4
&  3.4$\pm$1.3
&  3.5$\pm$1.4
&  2.9$\pm$1.1
& 3.3$\pm$0.8
\\
\hline
P\texttt{+}V &  4.3$\pm$0.7
&  4.1$\pm$0.8
&  4.4$\pm$0.6
&  4.1$\pm$0.7
& 3.6$\pm$1.1
\\
\hline
\multicolumn{6}{|c|}{\textbf{Customer C}}\\
\hline
Cond. & Fluen. & Trust & Alliance & Satis. & Effec.\\
\hline
G\texttt{+}N 
&  2.4$\pm$1.0
&  2.2$\pm$1.0
&  2.4$\pm$1.2
&  2.2$\pm$1.2
& 3.2$\pm$1.5
\\
\hline
P\texttt{+}N
&  2.8$\pm$1.1
&  2.6$\pm$1.1
&  2.7$\pm$1.1
&  2.9$\pm$1.7
& 3.1$\pm$1.5
\\
\hline
G\texttt{+}V
&  2.2$\pm$1.2
&  2.7$\pm$1.5
&  2.9$\pm$1.5
&  2.8$\pm$1.2
& 2.9$\pm$1.2
\\
\hline
P\texttt{+}V &  3.5$\pm$0.9
&  3.4$\pm$0.9
&  3.7$\pm$0.9
&  3.5$\pm$1.0
& 3.3$\pm$1.3
\\
\hline
\end{tabular}
\end{table}

Regarding the interaction experience reported in Table~\ref{tab:colab}, \textbf{visualisation} significantly impacted collaboration \textit{fluency} ($F(1,228) = 12.68$, $p << 0.001$), human-robot \textit{trust} ($F(1,228) = 19.13$, $p << 0.001$), working \textit{alliance} ($F(1,348) = 29.35$, $p << 0.001$), service \textit{satisfaction} ($F(1,108) = 4.81$, $p = 0.03$), and perceived service \textit{efficiency} ($F(1,108) = 4.71$, $p = 0.03$); Stopping \textbf{location} showed significant influence on \textit{fluency} ($F(1,228) = 16.89$, $p << 0.001$), \textit{trust} ($F(1,228) = 4.66$, $p = 0.03$), \textit{alliance} ($F(1,348) = 7.96$, $p = 0.01$), and \textit{satisfaction} ($F(1,108) = 12.99$, $p << 0.001$); \textbf{Customer} role showed significant influence on \textit{fluency} ($F(2,228) = 10.08$, $p << 0.001$), \textit{trust} ($F(2,228) = 10.46$, $p << 0.001$), \textit{alliance} ($F(2,348) = 19.65$, $p << 0.001$), and perceived service \textit{efficiency} ($F(2,108) = 5.81$, $p = 0.004$); The interaction between \textbf{visualisation} and stopping \textbf{location} showed significant influence on \textit{fluency} ($F(1,228) = 7.21$, $p = 0.01$); The interaction between stopping \textbf{location} and \textbf{customer} role showed significant influence on working \textit{alliance} ($F(2,348) = 3.30$, $p = 0.04$); The interaction between \textbf{visualisation}, stopping \textbf{location}, and \textbf{customer} role showed significant influence on working \textit{alliance} ($F(2,348) = 3.47$, $p = 0.03$).

We analysed the effect sizes of our three-way mixed ANOVA calculated as $\eta_{p}^{2}$. \textbf{Visualisation} has a moderate effect on \textit{trust} ($\eta_{p}^{2} = 0.08$) and working \textit{alliance} ($\eta_{p}^{2} = 0.08$); \textbf{Stopping location} has a moderate effect on perceived \textit{intelligence} ($\eta_{p}^{2} = 0.06$), collaboration \textit{fluency} ($\eta_{p}^{2} = 0.07$), and service \textit{satisfaction} ($\eta_{p}^{2} = 0.11$); \textbf{Customer role} has a moderate effect on collaboration \textit{fluency} ($\eta_{p}^{2} = 0.08$), trust ($\eta_{p}^{2} = 0.08$), and perceived service \textit{efficiency} ($\eta_{p}^{2} = 0.10$).

\subsection{Preference ranking}
After all the interaction sessions had been completed, the participants ranked their preference towards the four conditions from most to least preferred. In Table~\ref{tab:pref}, we report the percentage of each condition being chosen as the first preference for each customer role and for all participants. A Friedman test was conducted on the preference rankings and we found significant differences between the conditions for all participants and for all customer roles (all participants: $\chi^2 = 37.72$, $p << 0.001$; Customer A: $\chi^2 = 13.80$, $p = 0.003$; Customer B: $\chi^2 = 11.16$, $p = 0.01$; Customer C: $\chi^2 = 16.56$, $p = 0.001$). Personal shopping with visualisation was consistently ranked as the most preferred condition.

\begin{table}[htb]
\centering
\caption{Each condition being chosen as the first preference by customer role (A/B/C) and for all participants (\%). Visualisation with personalised stopping location was found to be the most preferred condition.}
\label{tab:pref}
\begin{tabular}{|l|r|r|r|r|}
\hline
Condition & A & B & C & All\\
\hline
G\texttt{+}N &  10\%&  30\%&  0\%& 13\%\\
\hline
P\texttt{+}N&  0\%&  0\%&  0\%& 0\%\\
\hline
G\texttt{+}V&  20\%&  30\%&  30\%& 27\%\\
\hline
P\texttt{+}V &  \textbf{70\%}&  \textbf{40\%}&  \textbf{70\%}& \textbf{60\%}\\
\hline
\end{tabular}
\end{table}
\subsection{Qualitative results}
From the free-text questionnaire responses, six participants (three Customer As, one B, two Cs) reported mild confusion in terms of left and right in the shown image and on the robot's tray in at least one of the interaction sessions. In terms of the stopping location, one Customer B commented the general stopping to be ``unlively'', two Customer Cs commented the robot ``felt kinda ignorant and uncaring for the customers'' and ``kinda rude'', while two Customer Bs considered general stopping to be ``much faster'' than personal stopping. One Customer A commented that personal stopping is ``much more pleasant'' than general stopping.

In the exit interviews, seven out of the ten groups reported the robot being ``smooth'', ``efficient'', ``fluid'' or ``friendly'', while three groups considered the robot ``detached'' or ``artificial''. Two groups reported preferring personal stopping, while two groups preferred general stopping as it felt faster and more efficient. Five groups reported that visualisation has benefited the interaction. In addition, three groups discussed desires for the robot to incorporate user inputs and adjust the interaction accordingly. Comparing the interaction with previous experiences with restaurant robots, two groups reported a positive impression as the robot was able to autonomously navigate between the kitchen and the table, while in their previous experience, the robots required human supervision or were followed by human staff. Moreover, these two groups reported the interaction was more efficient and smooth compared to their previous experiences. When hypothesising how the robot functioned, nine our of ten groups identified correctly that the robot followed pre-programmed waypoints with autonomous navigation and obstacle avoidance, while one group assumed the robot had used an onboard camera to recognise and locate humans during the interaction.

\section{Discussion}
\subsection{Major findings and implications}
This study demonstrates the influence of a robot's visualisation and motion on the objective and subjective service outcomes in a group order delivery scenario. Our work indicates that group service interaction introduces additional complexity as opposed to dyadic HRI often studied in previous research. Each customer in the same group can experience the interaction differently. Thus, instead of serving the group as a whole, personalised visualisation, robot-approaching behaviour and delivery location are recommended for an accurate order delivery and positive user experience.

Our results confirmed our first hypothesis that a robot displaying visualisation for communicating its intent has positive effects on the objective and subjective outcomes of group service interactions. This aligns with existing literature on visual displays for intent communication as discussed in Section~\ref{subsec:bg-vis}.

Regarding stopping location, delivering orders to individual customers (personal stopping) as opposed to delivering for the whole group together at a general location by itself is not a significant factor influencing order delivery accuracy. However, the interaction between visualisation and stopping location, as well as the interaction between the customer role (where a person is seated at the table in the group) and stopping location have a significant impact. Combining personal stopping location and visualisation results in the highest delivery accuracy and was the most preferred condition. In terms of subjective perception, stopping location had a significant influence on the perceived animacy, likeability, and intelligence of the robot, as well as the collaboration fluency and user satisfaction, with personal stopping yielding more positive perception than general stopping. Thus, our second hypothesis is supported. This extends current understanding with regards to the motion and physical proximity in HRI as discussed in Section~\ref{subsec:bg-motion}.

In this group interaction scenario, we found significant differences between the participants assuming each customer role on all measures of subjective perception towards the robot and the service interaction. While customer role by itself is not a significant factor influencing order delivery accuracy, the interaction between customer role and the robot's stopping location had a significant influence on the order delivery accuracy. Thus, our third hypothesis is supported.

Our study indicates that to improve the objective and subjective outcomes of a restaurant delivery robot when serving a group of customers, it is more beneficial to consider each customer individually as opposed to current practices of serving the group as a whole. Visualisation combined with personalised motions will facilitate a more accurate robot-to-human intention communication, as well as yield more positive interaction experiences and user perceptions. Such personalised service can contribute to mitigating robot and service failures in addition to existing approaches, such as apology and acknowledgement of failure discussed in Section~\ref{subsec:bg-delivery}, to improve robot waiters' outcomes in restaurant services.

\subsection{Limitations and future work}
Our experiments have been conducted in a controlled lab environment simulating the restaurant delivery service scenario. In the next step, we plan to extend this work to test the different robot intention communication approaches in the restaurant environment with an extended number and diversity in the participant population. In addition, as reported by some participants in the free text response and exit interviews, the visualisation can be further improved to reduce confusion, such as including LED displays on the robot's tray for indicating the order for a specific customer. Furthermore, we plan to extend the interaction design to allow the incorporation of user inputs to further personalise the interaction experience. The service interaction can be further improved with more engaging and expressive robot behaviours, such as including emotional expressions.

\section{Conclusion}
We investigate a social robot's intentional communication when interacting with a group of users in the restaurant order delivery context. Our study demonstrates the importance of combining movement and visualisation in a robot's behaviours for more interpretable and satisfying service interaction. Personalised movement trajectory and order delivery location combined with visualisation of the order information resulted in more accurate communication from the robot to its intended order receivers and a more positive user experience. Our findings demonstrate how a service robot's behaviours can be adapted to an individual customer's context and needs to yield better objective and subjective outcomes. Further, video recordings provided in the accompanying public dataset can benefit future research on robot-group interaction and better design of social robots for service scenarios.

\bibliographystyle{unsrt} 
\bibliography{ref}

\end{document}